\documentclass[11pt,a4paper]{article}
\usepackage{times,latexsym}
\usepackage{url}
\usepackage[T1]{fontenc}

\usepackage{gb4e}
\noautomath

\usepackage[acceptedWithA]{tacl2021v1}

\usepackage{pifont}

\usepackage{tikz-dependency}
\usepackage{bm}
\usepackage{booktabs} 
\usepackage{makecell}
\usetikzlibrary{positioning}

\usepackage{footnote}
\makesavenoteenv{tabular}
\makesavenoteenv{table}
\usepackage{comment}
\usepackage{graphicx}
\usepackage{enumitem}
\graphicspath{ {./figures/} }
\usepackage{gb4e}\noautomath
\usepackage{subcaption,siunitx,booktabs}

\usepackage{xspace,mfirstuc,tabulary}
\newcommand{\dateOfLastUpdate}{Dec. 15, 2021}
\newcommand{\styleFileVersion}{tacl2021v1}

\newif\iftaclinstructions
\taclinstructionsfalse 
\iftaclinstructions
\renewcommand{\confidential}{}
\renewcommand{\anonsubtext}{(No author info supplied here, for consistency with
TACL-submission anonymization requirements)}
\newcommand{\instr}
\fi

\iftaclpubformat 

\newcommand{\TaclPapers}{Final Versions\xspace}
\else

\newcommand{\TaclPapers}{Submissions\xspace}
\fi

\title{Formatting Instructions for TACL \TaclPapers \\
(Base files: \styleFileVersion-template.tex \& \styleFileVersion.sty, dated \dateOfLastUpdate)}

\author{Bingzhi Li \and Guillaume Wisniewski \and Beno\^it Crabb\'e\\
  Université Paris Cité, LLF, CNRS, 75\,013 Paris, France \\
  {\tt bingzhi.li@etu.u-paris.fr} \\
  {\tt \{guillaume.wisniewski,benoit.crabbe\}@u-paris.fr} \\}

\date{}

\title{Assessing the Capacity of Transformer to Abstract Syntactic Representations: A Contrastive Analysis Based on Long-distance Agreement}

\begin{document}
\maketitle
\begin{abstract}
  Many works have shown that transformers are able to predict
  subject-verb agreement, demonstrating their ability to uncover an
  abstract representation of the sentence in an unsupervised way.
  Recently, \newcite{li-etal-2021-transformers} found that
  transformers were also able to predict the object-past participle
  agreement in French, the modeling of which in formal grammar is
  fundamentally different from that of subject-verb agreement and
  relies on a movement and an anaphora resolution.

  To better understand transformers internal working, we propose to
  contrast how they handle these two kinds of
  agreement. Using probing and counterfactual analysis methods, our
  experiments on French agreements show that \textit{i)} the agreement
  task suffers from several confounders which partially question the
  conclusions drawn so far and \textit{ii)} transformers handle
  subject-verb and object-past participle agreements in a way that is
  consistent with their modeling in theoretical linguistics.

\end{abstract}
 
\section{Introduction}

Since~\newcite{linzen-etal-2016-assessing}, the long distance
agreement task has been a paradigmatic
test for assessing the ability of Neural Network Language Models (NNLMs) to
uncover syntactic information from raw texts: a model able to predict
the correct verb form~(especially when the verb does not agree with
the word just before it) has to, to some extent, acquire a
representation of the syntactic structure and encode it in its
internal representations.
 
In this work, we seek to identify to which extent the NNLM abstracts
its representations from surface pattern recognition to structurally
motivated representations. To do this, we focus on two kinds of number
agreement in French (both morphologically marked):
\begin{exe}
   \ex\label{ex:agrS}
   \gll Les \textbf{chat·s} [ que Noûr aime bien
  ]\textcolor{blue}{$_{RC}$} \textbf{jou·ent} dans le jardin.   \\
   {\tiny The\_Pl} {\tiny cats\_\textbf{Pl}} {\tiny [} {\tiny that} {\tiny Noûr} {\tiny likes\_Sg} {\tiny a\_lot} {\tiny ]$_{RC}$} {\tiny play\_\textbf{Pl}} {\tiny in} {\tiny the } {\tiny garden.} \\
   \ex\label{ex:agrO}
  \gll Il aime les \textbf{chat·s} [ que Noûr a \textbf{adopté·s} ]\textcolor{blue}{$_{RC}$}. \\
   {\tiny He\_Sg} {\tiny loves\_Sg} {\tiny the\_Pl} {\tiny cats\_\textbf{Pl}} {\tiny [} {\tiny that} {\tiny Noûr} {\tiny has\_Sg} {\tiny adopted\_\textbf{Pl}} {\tiny ]$_{RC}$} \\
\end{exe}
These sentences involve two agreements: the first one between a noun
\textit{chats} and the main verb \textit{jouent} and the second one
between the same noun and the past participle \textit{adoptés}.\footnote{
  The two agreements could occur in the same sentence like in the
  example of Figure~\ref{fig:ex_agreement}.} A naive look at
(\ref{ex:agrS}) and (\ref{ex:agrO}) may suggest that they are two
identical agreements between a noun and a verbal form separated by a
few words.  Yet from a linguistic perspective these two agreements
receive a substantially different analysis. As in English, example
(\ref{ex:agrS}) involves a number agreement between the main clause
verb and its subject where an embedded relative clause occurs between
the two. To predict this kind of agreement, the model has to learn an
abstract representation that is not solely based on the linear
sequence of words but also on the long syntactic dependency between
the verb and its subject, in order to ignore the linear proximity of
the noun~\textit{Noûr}.

On the other hand, the agreement between \textit{chats} and
\textit{adoptés} in (\ref{ex:agrO}) is an object relative clause
agreement. Overall, it involves an agreement between a noun in the
main clause and a past participle in the relative clause. To predict
this kind of agreement, the model must be able to detect a complex set
of patterns across different clauses that brings into play an anaphora
resolution and a movement, a set of operations whose nature is
fundamentally different from phrase structure embedding involved in
the subject-verb agreement in (\ref{ex:agrS}).


While these two kinds of agreement show very similar surface patterns,
their modelings in formal grammar result in completely different
representations of the sentence structure, and it is not clear whether
and how a sequential language model can identify these abstract
representations based merely on the words sequence. Even though a
tremendous number of studies have brought to light the ability of
neural networks to capture and abstract the information needed to
predict the subject-verb agreement (see Section~\ref{sec:rw} for an
overview), it is only very recently that
\newcite{li-etal-2021-transformers} showed that they were also able
to predict the much rarer past participle agreement across clause
boundaries like the agreement between \textit{chats} and
\textit{adoptés} in (\ref{ex:agrO}).

Building on  \newcite{li-etal-2021-transformers,li-etal-2022-distributed}, the goal of the
present work is to contrast how transformers handle these two kinds of
agreement: we aim to determine whether they encode the \emph{same}
abstract structure in their internal representations to capture the
information required to choose the correct verb form or, on the
contrary, if the abstract structure they encode reflects the
\emph{distinction} made in the theoretical modeling of these two
agreements. This contrast will shed a new light on our understanding
of the internal working of transformers.

The contributions of this paper are twofold.  First, we assess
incremental transformers syntactic ability to process two different
syntax-sensitive dependencies with similar surface forms in
French. Our results show that transformers perform consistently well
on both agreement phenomena, and crucially, that they are able to
abstract away from potential confounds such as lexical co-occurences
or superficial heuristics.  Second, we use linguistic probes as well
as targeted masking intervention on self-attention to test
\textbf{where} transformer-based models encode syntactic information
in their internal representations and \textbf{how} they use it. We
find that, for both constructions, even though the long-distance
agreement information is mainly encoded locally across the tokens
between the two elements involved in the agreement, transformers are
able to leverage distinct linguistically motivated strategies to
process these two phenomena.\footnote{ Datasets and code are available at \url{https://github.com/bingzhilee/contrastive_analysis}}

The rest of this paper is organized as follows: First, in
Section~\ref{sec:long-distance}, we define more precisely the two
kinds of agreement considered in this work. Then, in
Section~\ref{sec:number_agreement_task}, we contrast the capacity of
transformers to capture subject-verb and object-past participle
agreement. Following this, in Section~\ref{sec:probing_causal}, by using probing
and counterfactual analysis, we examine the way transformers encode
and use the syntactic information to process these two linguistic
phenomena. We then discuss, in Section~\ref{sec:disc}, the impact of some potential
confounders, in Section~\ref{sec:rw} the related work and finally we conclude in Section~\ref{sec:conclu}.

\section{Long-distance Agreement\label{sec:long-distance}}

\begin{figure*}[htbp]
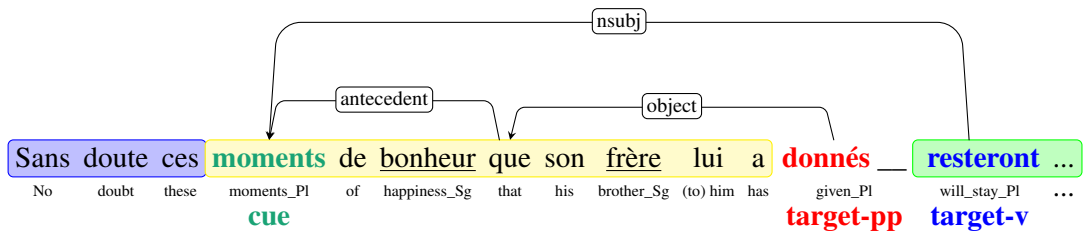

  \centering
  \begin{dependency}
    \tikzstyle{POS}=[font=\tiny]
    \depstyle{pref}{draw=blue, fill=blue!25}
    \depstyle{cont}{draw=yellow, fill=yellow!25}
    \depstyle{suf}{draw=green, fill=green!25}
    
    \begin{deptext}
    Sans \& doute \& ces \& \textcolor{blue!40!green!90}{\bf moments}\& de \& \underline{bonheur} \&  que \& son \& \underline{frère} \& lui \& a \& \textcolor{red}{\bf donnés}  \_\_ \& \  \textcolor{blue}{\bf resteront} \& ... \\
        |[POS]| No \& |[POS]| doubt \& |[POS]| these \& |[POS]| moments\_Pl \& |[POS]| of \& |[POS]| happiness\_Sg \& |[POS]| that \& |[POS]| his \& |[POS]| brother\_Sg \& |[POS]| (to) him \& |[POS]| has  \& |[POS]| given\_Pl \& |[POS]| will\_stay\_Pl \& ...\\ 
     \&  \&  \& \textcolor{blue!40!green!90}{\bf cue} \&  \&  \&  \& \&  \& \& \& \textcolor{red}{\bf target-pp}  \& \textcolor{blue}{\bf target-v} \&  \\
    \end{deptext}

    \depedge[edge unit distance=1ex]{7}{4}{antecedent}
    \depedge[edge unit distance=.5ex]{12}{7}{object}
    \depedge[edge unit distance=1ex]{13}{4}{nsubj}
    \wordgroup[pref]{1}{1}{3}{prefix}
    \wordgroup[cont]{1}{4}{11}{context}
    \wordgroup[suf]{1}{13}{14}{suffix}
  \end{dependency}
  \caption{Example of object-past participle agreement and long-distance subject-verb agreement. To predict the target past participle (in red) number, a human is expected to get the feature from the object relative pronoun \textit{(que)} that gets it from its antecedent (\textit{moments} in bold green).  The latter is also the grammatical subject of the main verb (\textit{resteront} in bold blue) and determines its number. For the object-pp agreement, the prefix is highlighted in blue, the context in yellow and the suffix in green. 
  \label{fig:ex_agreement}}
\end{figure*}

Overall, this work aims to assess to which extent transformers rely on
shallow surface heuristics to predict agreement or to which extent
they infer more abstract knowledge. To study this question, we rely on
a contrast between two agreement tasks as exemplified in
(\ref{ex:agrS}), (\ref{ex:agrO}) and in Figure~\ref{fig:ex_agreement}. Both involve a
long-distance agreement but they receive substantially different
linguistic analyses.

\paragraph{Syntactic phenomena} In this work, we focus on corpus-extracted sentences
involving \emph{object relatives} such as the one analyzed in
Figure~\ref{fig:ex_agreement}. These kinds of sentences could  
involve two types of agreement, which receive
substantially different linguistic analyses. The first one between the
“cue” and “target-v” is an agreement between a verb and its subject
separated by an object relative clause. As exemplified in 
Figure~\ref{fig:ex_agreement}, the number of the main verb
\textit{resteront} (will\_stay Plural) is determined by the head of the subject
\textit{moments}. In this first agreement, there is no relevant
relationship between the relative pronoun \textit{que} and the main verb. To correctly
predict the number of the verb, a model must infer an abstract
structural dependency across the relative clause to distinguish the embedded subject (\textit{frère}) from the main clause subject (\textit{moments}). 
The model has to resist
the lure of the linearly closer but irrelevant attractors \textit{frère} and \textit{bonheur} (attractors, underlined in Figure~\ref{fig:ex_agreement}, are intervening nouns with misleading agreement
features).

We also consider the agreement of the past participle in the relative clause:  the past-participle (denoted
“target-pp” in Figure~\ref{fig:ex_agreement}) agrees with its complement (the “cue”) if the latter moves
before it. This agreement relies on an abstract set of relations
between tokens occurring in different clauses. It involves an anaphora (designated by the \texttt{antecedent} arc)
and a filler-gap dependency: the filler is \textit{que} and the gap, indicated with a line in Figure~\ref{fig:ex_agreement}, is an empty syntactic position licensed by the filler. The relative pronoun
\textit{que} is the pre-verbal direct object of the past participle
\textit{donnés} and triggers the agreement of the
past participle. To obtain its agreement features, the relative
pronoun has to be linked by anaphora to its nominal antecedent
\textit{moments}. In other words, to correctly agree the past participle
in theory, it is necessary to identify the object relative pronoun and
its antecedent. The model has also to ignore the effect of attractors
occurring between the antecedent of the relative pronoun and the past
participle. 

\paragraph{Terminology} In this paper, for both agreement phenomena, we refer to the noun item as
the \textbf{cue}, and the verbal item as
the \textbf{target} (see Figure~\ref{fig:ex_agreement}). \emph{pp} is short for past participle. We call the
tokens before the cue the \emph{prefix}, the tokens between the cue
and the target the \emph{context}, and the tokens after the target the
\emph{suffix}. We only consider number agreement as \textit{i)} number
agreement is the only feature shared by the two agreements we
consider\footnote{In French, the verb has to agree in number with its
  subject, and the past participle conjugated with the auxiliary
  \textit{avoir} agrees in number \emph{and} in gender with its direct object
  if the latter appears before it.}  \textit{ii)} the main purpose is to
design reasonably simple patterns allowing to extract a
sufficiently large number of representative examples.

\paragraph{Datasets} For object-past participle agreement, we consider
the number agreement evaluation set introduced by
\newcite{li-etal-2021-transformers}: to study the ability of
transformers to predict the object-past participle agreement, they
have parsed automatically the French Gutenberg Corpus, and have
extracted, with simple heuristics a set of 68,497 French sentences
(65\% singular and 35\% plural) involving an object relative. For the
subject-verb agreement considered in this paper, we extracted from the
same parsed corpus 27,582 sentences (70\% singular 30\% plural), in
which the sequence between the subject and the verb contains at least
one object relative clause. There are less items in subject-verb agreement evaluation set because noun phrases modified by relative clause(s)  occur more frequently in the object position of the main clause as in example (\ref{ex:agrO}) than in the subject position like in example (\ref{ex:agrS}). In these two
evaluation sets, an arbitrary number of words can occur between the
\emph{cue} and the \emph{target}: in average, there are 5 tokens
between the antecedent and the past participle, 11 tokens between the
subject and the verb. The intervening tokens include varied
constructions such as prepositional phrases, participials or nested
relative clauses, which would make the agreement tasks more
challenging. We also ensure that the two evaluation sets are
completely separate from the training data of LMs.

\section{Number Agreement Prediction}\label{sec:number_agreement_task}

\begin{table*}
  \scalebox{.9}{
    \begin{tabular}{ccl}
      \toprule
      \makecell{Number of \\ heuristics} & \makecell{Difficulty \\ of agreement} & Examples \\
      \midrule
      5 & \texttt{-{}-{}-} & Si les \textbf{idées} que ces mots représentent ne \textbf{sont} pas ...\\
                   & & {\tiny $_{\textrm{\textcolor{orange}{(5)}}}$\textit{If the ideas$_{\textrm{\textcolor{orange}{(4)}}}^{\textrm{\textcolor{orange}{(1)}}}$ that these words$_{\textrm{\textcolor{orange}{(2)}}}$ represent$_{\textrm{\textcolor{orange}{(3)}}}$  are not...}}\\
      4 & \texttt{-{}-} &  Les \textbf{choses} que nous avions vues cent fois avec indifférence nous \textbf{touchent}...\\
                  && {\tiny $_{\textrm{\textcolor{orange}{(5)}}}$\textit{
      The things$_{\textrm{\textcolor{orange}{(4)}}}^{\textrm{\textcolor{orange}{(1)}}}$  that we had seen a hundred times with indifference us$_{\textrm{\textcolor{orange}{(3)}}}$ touch }}\\
    3 & \texttt{-} & Un philosophe est curieux de savoir si les \textbf{idées} qu' il a semées \textbf{auront}... \\
                &&  \textit{\tiny A philosopher is curious to know if the ideas$_{\textrm{\textcolor{orange}{(2)}}}^{\textrm{\textcolor{orange}{(4)}}}$ that he has sown$_{\textrm{\textcolor{orange}{(3)}}}$ have... }     \\

     2 & \texttt{+} &Les \textbf{emblèmes} qu' on y rencontre à chaque pas \textbf{disent} ...  \\
                 && \textit{\tiny The emblems$_{\textrm{\textcolor{orange}{(4)}}}^{\textrm{\textcolor{orange}{(1)}}}$ that we meet at each step say ... }    \\     
    1 & \texttt{++} & Les \textbf{qualités} qui t'ont fait arriver si jeune au grade que tu as \textbf{doivent} te porter ... \\ 
               && \textit{\tiny The qualities$_{\textrm{\textcolor{orange}{(1)}}}$ that made you arrive so young at the rank you have must bring you ...}\\

    0 & \texttt{+++} & Ce soir les \textbf{hommes} que j'ai postés sur la route que doit suivre le roi \textbf{prendront} ...    \\
                                         && \textit{\tiny Tonight the men that I have posted on the road that the king must follow  will\_take ... }\\
      \bottomrule
    \end{tabular}}
  \caption{Examples from our evaluation set of subject-verb agreement, stratified by the count of
    surface heuristics predicting the \emph{target}'s number, a proxy
    to the task difficulty. The target verbs and their subjects are in
    bold. Orange numbers in parenthesis indicate the presence of different types of heuristics   \label{tab:examples_h} }
\end{table*}

\subsection{Experimental Setting}

Following~\newcite{linzen-etal-2016-assessing}
and~\newcite{gulordava-etal-2018-colorless}, we use the number
agreement task to evaluate incremental transformers' ability to capture syntactic information: given a sentence,
the model is fed with all tokens preceding the target verb (either the
verb of the main clause or the past participle in the relative clause)
and we compare the probabilities it assigned to the singular and plural
variants of the target. The model's syntactic ability is measured by the
percentage of sentences for which the verb form with the higher
probability is the one that respects the agreement rules of the
language (i.e.\ matches the number of the \emph{cue}).

To measure the prediction difficulty of the evaluation sentences in the two agreement tasks, we follow
\newcite{li-etal-2021-transformers} and define a scale that quantifies
the agreement difficulty for a given sentence by counting the number
of surface heuristics that predict the correct form of the target
verb. We target five superficial heuristics, namely (1) the \emph{first noun} of the sentence, (2) the \emph{closest noun}, (3) the \emph{closest token} with a mark of number, (4) the \emph{noun} before the closest \emph{que} and (5) the \emph{majority number} expressed in the sequence preceding the \emph{target}: as illustrated in Table~\ref{tab:examples_h}, the correct form
of the target verb in the \textit{5 heuristics} subset (the “easiest”
subset of our evaluation sets) can be easily predicted by simply
memorizing any of the five surface heuristics (e.g. the target form should match the first noun in the sentence). On the contrary, the
\textit{0 heuristic} subset gathers sentences for which the target verb form can only be predicted by constructing an abstract representation
of the sentence. We will mainly focus on the hardest cases (i.e.\ \emph{0}
and \emph{1 heuristic} subsets) in our evaluation.

\paragraph{Models} The experiments were carried out with two
incremental language models designed by
\newcite{li-etal-2021-transformers}: an LSTM and an incremental
transformer language model. Both models perform predictions
incrementally using the conditional probability
$P(w_i|w_1\ldots w_{i-1};\boldsymbol\theta)$. The LSTM model has 2
layers and the transformer model has 16 layers and 16 heads. Word
embeddings are of size 768. Note that the LSTM model has less
parameters than the transformers and, consequently, a direct
comparison of their performance is not completely fair. As training an
LSTM with a comparable number of parameters is hardly computationally
tractable and LSTM language model has been shown to perform
consistently well across varied agreement dependencies
\cite{linzen-etal-2016-assessing,gulordava-etal-2018-colorless,marvin-linzen-2018-targeted},
we only use it in this work as a strong baseline model and focus, in
our analyses, on transformers. Hyper-parameters were chosen by
minimizing the perplexity on the validation set and the best
hyper-parameters were used to train for each architecture five models
on the French Wikipedia corpus (90 million tokens). All the results
reported in this paper are averaged across five models. More details
about the models are given in Section~\ref{sec:LM} in the appendix.

\subsection{Number Prediction Results}\label{subsec:number_pred_results}

\paragraph{Overall accuracy:} The two architectures are able to
predict both long-distance agreements with a very good
accuracy:\footnote{Detailed scores are provided in
  table~\ref{tab:original-heuristics} in the appendix.} transformers
made correct number prediction in 94.6\% of the object past participle
agreement (LSTM 82.1\%) and in 98.9\% of the subject-verb agreement
(LSTM 94.3\%), a result consistent with the conclusions drawn by
previous studies \cite{linzen2021syntactic}.

\begin{figure}[h]
\includegraphics[width=\columnwidth]{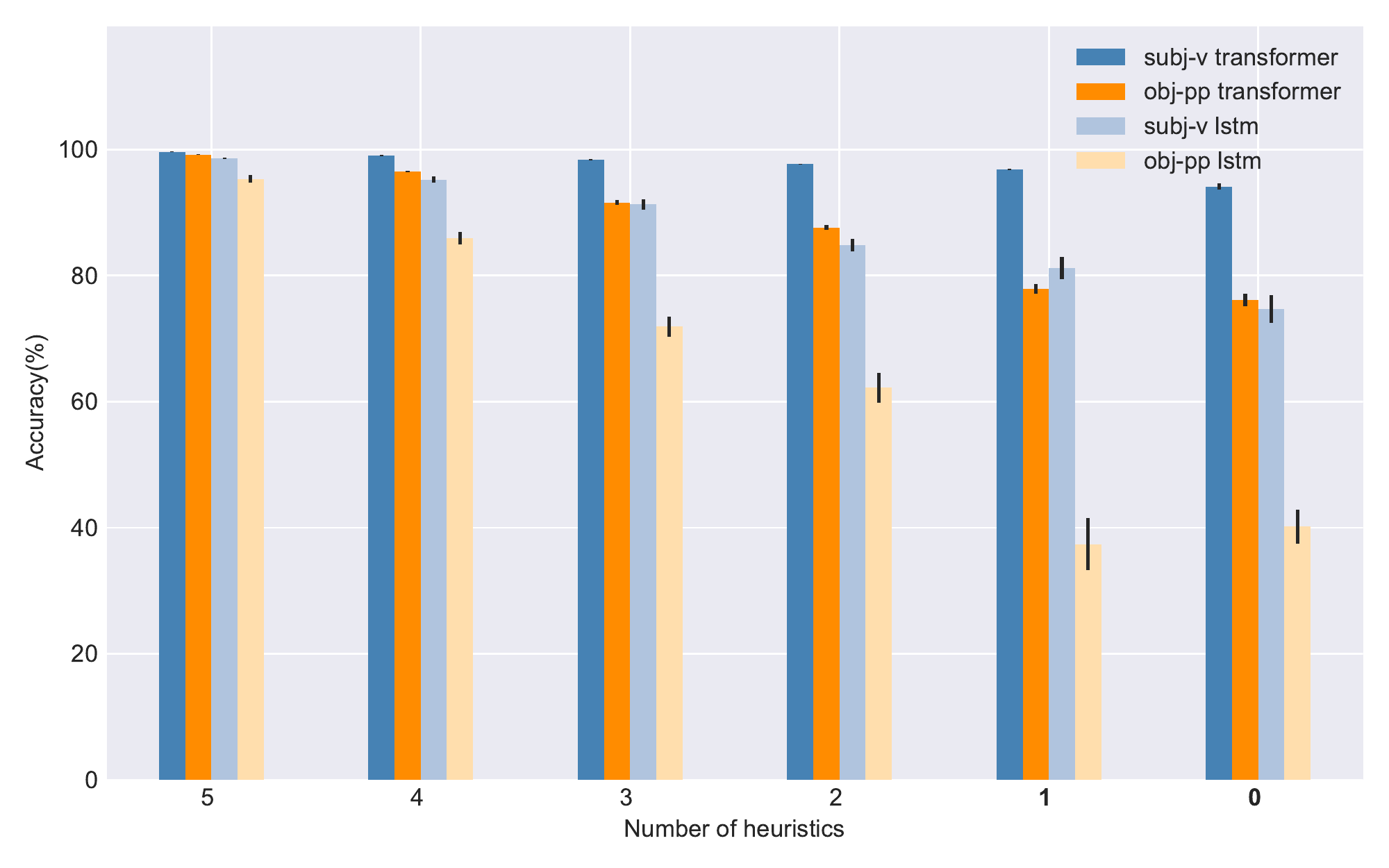}
\caption{Accuracies achieved by  transformers and LSTMs (indicated by lighter color bars) as a
  function of the agreement prediction difficulty. Blue bars represent the subject-verb agreement and orange bars for the object-pp agreement. The more heuristics are present, the easier the task is.  }\label{tab:original-heuristics_graph}
\end{figure}
 
\paragraph{Surface heuristics:}
However, detailed scores according to the difficulty of the task in Fig.~\ref{tab:original-heuristics_graph} show
more nuanced results. With respect to the type of agreement, we
observe that both LSTMs and transformers achieve much better
performance for subject-verb agreement than for the object-past
participle agreement, especially in the hardest cases (i.e.\ \emph{0 \& 1
heuristic} subsets), even though the linear distance between the \emph{cue}
and the \emph{target} in the subject-verb dependency is twice as long as that
in the object-past participle agreement (11 tokens vs. 5 tokens on
average).  This performance difference could result from the agreement
frequency in the training data: the subject-verb agreement exists
nearly in each sentence of the training data, while only 0.35\% of the
training sentences involve an object-past participle
agreement.

However, we do find a similar pattern across these two agreement
tasks: models' performances always decrease with the task difficulty. This observation generalizes
the conclusions of \newcite{li-etal-2021-transformers}: it shows that
the impact of surface heuristics is not limited to a relatively
infrequent and complex kind of agreement (i.e. the object-past
participle agreement) but also concerns the subject-verb agreement,
further confirming that results on long-distance agreement tasks
should be interpreted with great care.

It also appears that transformers outperform LSTM across the board
even though, as explained above, the performance of these two models are
not directly comparable: for the two types of agreements, transformers
are able to predict the correct verb form most of the time, even in the
hardest cases for which the performance of LSTM is below chance
level. Above all, this comparison highlights the very strong capacity
of transformers to capture syntactic information that even LSTM, a
very good baseline on which are based most of the conclusions drawn so
far on the syntactic capacity of neural networks, is unable to
capture.

\section{Do Transformers Process the Two Agreements in the Same Way?}\label{sec:probing_causal}

In the last section, transformer language model has demonstrated a
strong ability for capturing syntactic information that generalizes
beyond surface heuristics when processing long-range subject-verb and
object-past participle agreements. In this Section, we investigate
whether transformers acquire and use a meaningful, abstract
representation of the sentence to predict the two theoretically
distinct agreements we considered in this work or, on the contrary,
whether they use one unified agreement mechanism to resolve the two
agreement tasks. To do this, we use in Section~\ref{sec:probing}
linguistic probes~\cite{veldhoen2016diagnostic,conneau-etal-2018-cram}
to compare the distribution of the number information across the
sentence token representations. Then, in Section~\ref{sec:causal}, we
compare the way transformers use the encoded syntactic information to
process these two different long-range dependencies.

\subsection{Probing Contextualized Representations\label{sec:probing}}
We now investigate how transformers represent the syntactic
information required to predict the correct verb form: is it
represented across all the tokens following the \emph{cue} in the
sentence as made theoretically possible by the self-attention
mechanism and observed by~\newcite{klafka-ettinger-2020-spying}, or is
it encoded mainly locally around the \emph{cue} and the \emph{target}
tokens? To this end, we use linguistic probes. A probe is a classifier
trained to predict linguistic properties (in our case: the number of
the target verb) from the model representations (the representation of
a token uncovered by the model). Intuitively, a probe that achieves
high accuracy implies that these properties were encoded in the
representation. To test the two hypotheses and find out whether number
information is encoded only locally or globally in the whole sentence,
we simply have to train probes for all words in different parts of the
sentence.

More precisely, for each sentence of our evaluation set, we extract
the token representations from the last layer of the transformer and
associate them with a label describing the number of the target
verb. We then train one logistic regression classifier\footnote{All
  classifiers were implemented with the \texttt{scikit-Learn}
  library~\cite{scikit-learn}, setting the \texttt{max\_iter}
  parameter to 1,000 and the \texttt{class\_weight} to
  \texttt{balanced}.} for each PoS tag of each part of the sentence
(\emph{prefix}, \emph{context} or \emph{suffix} as defined in the
“Terminology” paragraph of Section~\ref{sec:long-distance}). We
consider 80\% of the examples as training data and use the remaining
20\% to evaluate the probe's performance.

\begin{table}
\scalebox{1}{
  \begin{tabular}{l cc cc c}
    \toprule 
    & \multicolumn{4}{c}{Mean probing Accuracy } \\
    \cline{2-5}
    & \makecell{Object-pp \\ } & \phantom{\tiny a} &
    \makecell{Subject-verb} & \phantom{\tiny a}  \\
    \cline{2-2} \cline{4-4} 
    \emph{prefix}        & 58.6\%$_{\pm 0.1}$ && 59.5\%$_{\pm 0.2}$ &\\
    \emph{context}       & 92.3\%$_{\pm 0.2}$ && 93.0\%$_{\pm 0.1}$ \\
    \emph{suffix}       & 73.6\%$_{\pm 0.2}$ && 78.1\%$_{\pm 0.2}$  \\
    \bottomrule 
  \end{tabular}}
\caption{Probing task results across different sentence parts (see
  Figure \ref{fig:ex_agreement}). Reported scores are the average of
  all the PoS-based classifiers' accuracies for each part of the
  sentence. \label{res:probing_high_level}}
\end{table}

\begin{figure*}[t]
    \centering\includegraphics[width=.49\textwidth]{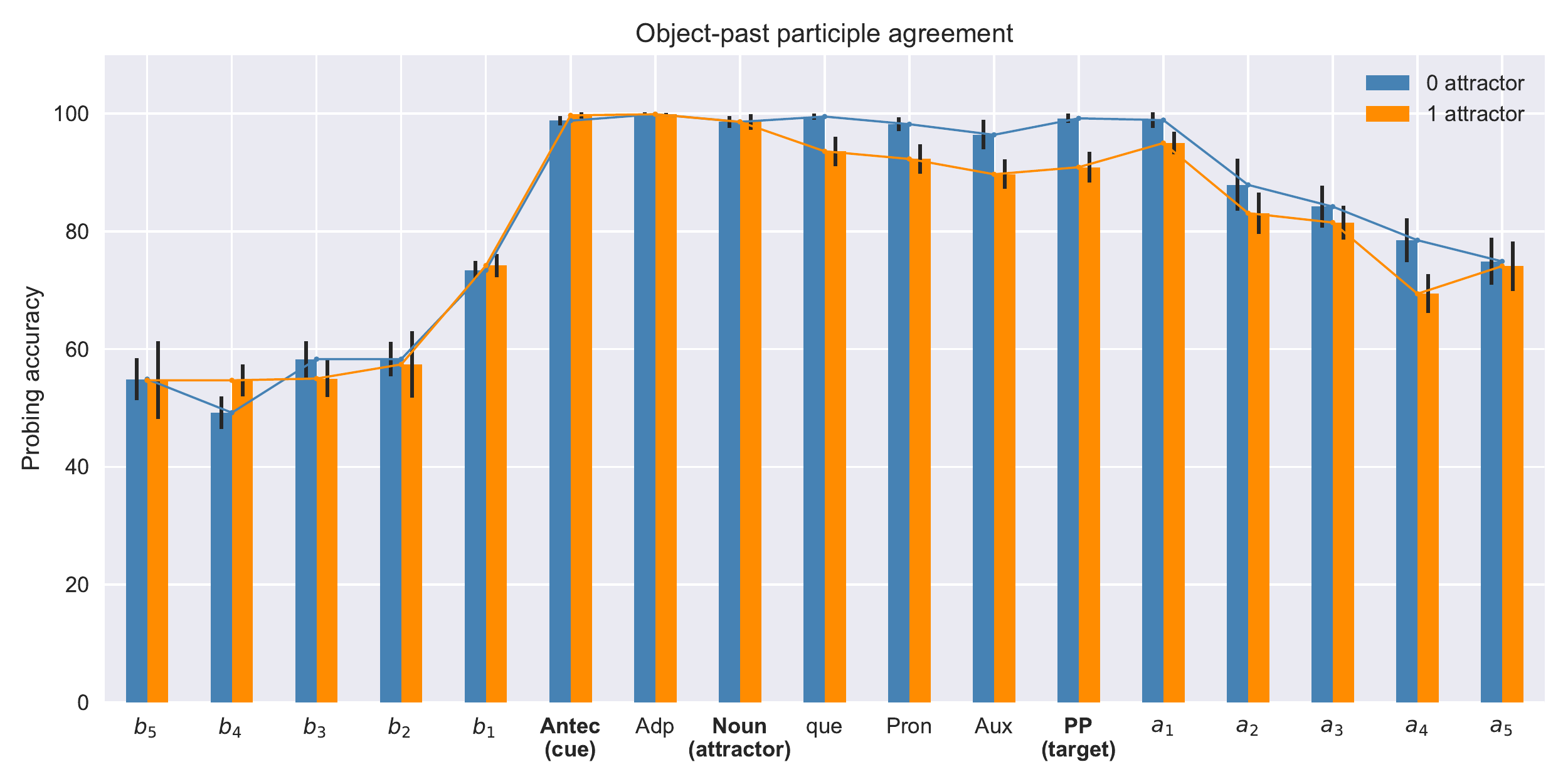}
    \centering\includegraphics[width=.49\textwidth]{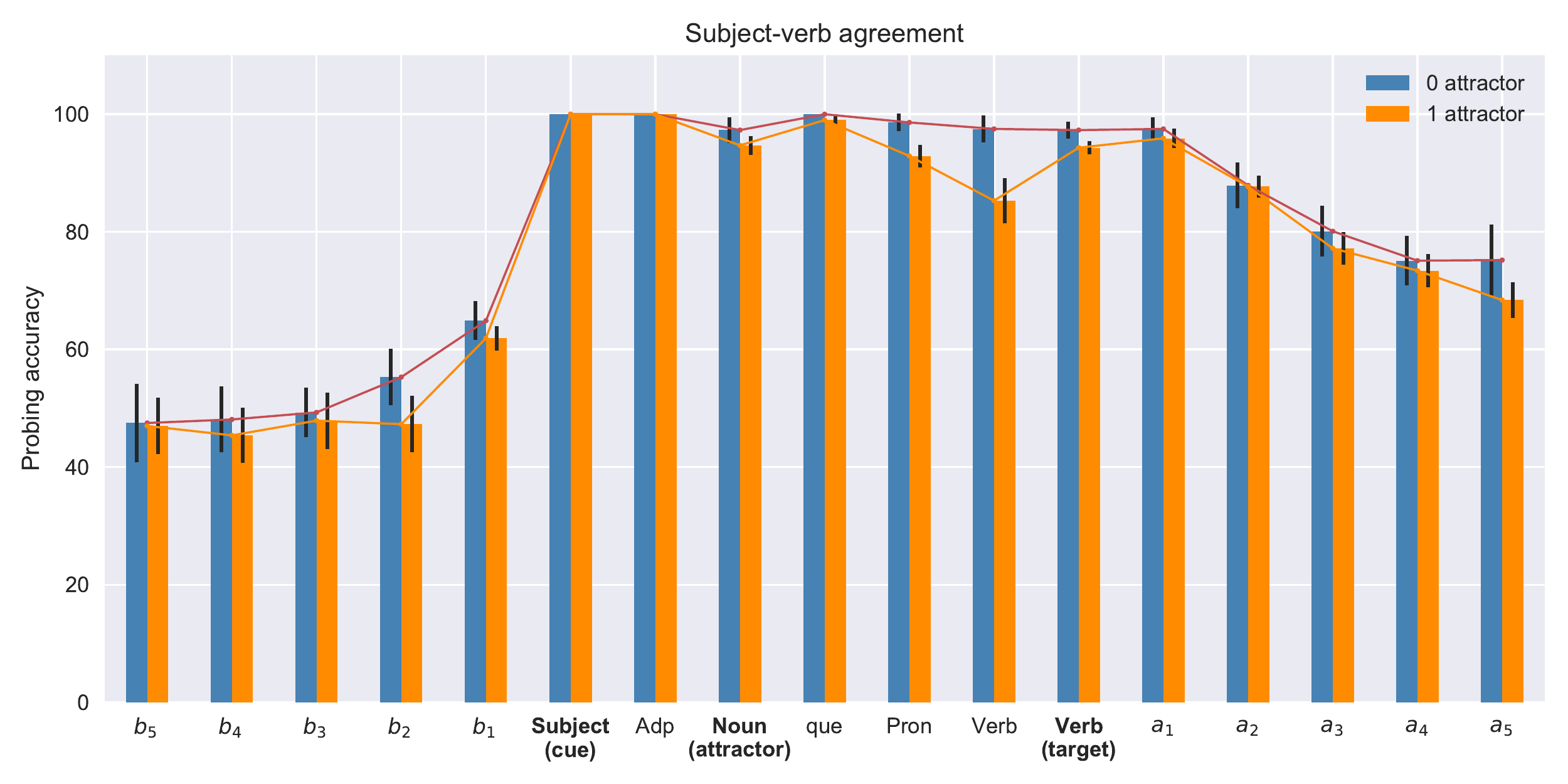}
    \caption{Probing accuracy at each position based on the number of the
      \emph{target}. The \texttt{bI} (resp. \texttt{aI}) position
      denotes the $I$-th token before (resp. after) the pattern. 
      \label{fig:fix5}}
\end{figure*}

\paragraph{Results} Table~\ref{res:probing_high_level} reports the
average accuracy achieved by our probes on different
parts of the sentence. We observe a similar pattern for the object-pp
and subj-verb agreement: the \emph{target} number information is
essentially encoded locally within the tokens of the
\emph{context} and is not represented uniformly across all the sentence tokens.

As expected, in the two cases, the performance of the probe on
\emph{prefix} is very low: since we consider an incremental model, the
tokens of the \emph{prefix} can not attend to the \emph{cue} and their
representation can not encode information about its number. The
accuracy mainly reflects the imbalance between singular and plural
forms in our evaluation set.

By contrast, the accuracy becomes consistently high when the tokens of
the \emph{context} are considered as input features, while accuracy on
the \emph{suffix} tokens drops sharply even though it remains better
than that observed in the \emph{prefix}, suggesting that the
information required to predict the correct \emph{target} form in both
dependencies is spread over all tokens between the \emph{cue} (where
the number of the \emph{target} verb is specified) and the
\emph{target} (where the information is “used”).

\paragraph{Representations within the context} 

To get a more accurate picture of how the number information is
distributed within the \emph{context}, we focus on a specific sentence
pattern: we only consider sentences for which the \emph{cue} is
separated from the relative pronoun only by a prepositional
phrase. For the object-past participle agreement, the pattern thus
corresponds to the sequence \texttt{Antecedent ADP N que
  PRON AUX} as in:
\begin{exe}
   \ex
   \gll ... \textbf{bureau·x} en métal qu' il a \textbf{trouvé·s}  \\
   ... {\tiny desks} {\tiny Prep.} {\tiny metal} {\tiny that} {\tiny he} {\tiny has} {\tiny found\_Pl}... \\
\end{exe}
For subject-verb agreement, it corresponds to the sequence \texttt{Subject ADP N que PRON V} as in:
\begin{exe}

    \ex
   \gll ... \textbf{bureaux} en métal qu' il aime \textbf{coût·ent} ... \\
   ... {\tiny desks} {\tiny Prep.} {\tiny metal} {\tiny that} {\tiny he} {\tiny loves} {\tiny cost}... \\
\end{exe}

We trained one probing classifier for each position on 800~randomly
sampled examples.\footnote{We used three sampling seed and for each
  sampling, three train/test splits. The sample is constrained to
  ensure balance between singulars and plurals in training and test
  sets.} We then evaluated the probe on a balanced set of 200
examples, differentiating sentences in which the embedded noun is an
attractor from sentences in which this noun has the same number as the
\emph{cue}.

Figure~\ref{fig:fix5} reports the probing accuracy at each
position. We observe that in the \emph{prefix} (i.e.\ b-positions) the
probe accuracy is low for the two kinds of agreement. On the contrary,
in the \emph{context}, the predictions of the probe are almost
perfect, even when there is an attractor. It is quite remarkable that,
in the \emph{suffix}, once the \emph{target} is encountered, the accuracy on the
subsequent tokens drops quickly, especially in the presence of an
attractor. These observations are
consistent across the two agreement tasks we consider.

The results of the probing experiments show us that transformer
language models encode the syntactic information in a very similar way
across the two long-range agreements. In terms of acquired
abstractions, nothing from the probing methodology allows to conclude
that the model acquires substantially different representations for
each agreement phenomenon.

\subsection{Causal Intervention on Attention\label{sec:causal}}

\begin{table*}[!ht]
  \centering
  \scalebox{1.0}{
  \begin{tabular}{l|ccccccc|cc}
    \toprule
    \phantom{ab}& <bos> & Les & cadeau\textbf{x} & que& le & directeur& a & accepté·\textbf{s} / accepté* \\
    \phantom{ab}&  & \scriptsize The\_Pl & \scriptsize gifts\_\bf{Pl} & \scriptsize that & \scriptsize the & \scriptsize director & \scriptsize has & \scriptsize accepted\_\textbf{Pl} / \scriptsize accepted\_Sg*\\
    \midrule
    Original& &-2.8&-9.5 &-7.3&-1.8&-6.1&-3.9&\textbf{-5.9} / -8.3\\
    Mask `que' &&-2.8&-9.5 &-7.3&-1.8&-6.1&-3.9&-13.7 / \textbf{-11.9} \\

    \bottomrule
  \end{tabular}}
\caption{Example sentence processed by our transformer LM, without intervention and with masking `que' intervention. We report the
  log-probabilities for each token of the sentence prefixes containing either the
  plural form of the target verb \textit{acceptés}, or its singular
  form \textit{accepté}. \label{ex:mask_que} }
\end{table*}

\begin{table*}
  \centering
  \scalebox{0.85}{
  \begin{tabular}{lrccccc}
    \toprule
    
    \multicolumn{7}{l}{\textit{Object-past participle}} \\
    Subsets & \makecell{Size \\ {\scriptsize (in sentences)}} & Original&  \makecell{Mask \emph{context} tokens  \\ except \texttt{cue que} }  & Mask \texttt{cue} &Mask \texttt{que}& Mask \texttt{cue+que} \\
    \midrule
    
    \phantom{ab} Overall      & 68,497 & 94.6 $_{\pm 0.4}$ &87.1$_{\pm 1.0}$ & 71.4$_{\pm 0.8}$  &78.8$_{\pm 0.5}$ &66.9$_{\pm 0.3}$ \\\cline{1-7}
    \phantom{ab} 5 heuristics  & 32,149 & 99.3 $_{\pm 0.05}$ &98.1$_{\pm 0.1}$& 93.4 $_{\pm 0.3}$&95.9$_{\pm 0.2}$ & 92.5$_{\pm 0.5}$\\
    \phantom{ab} 4 heuristics & 12,711 & 96.3 $_{\pm 0.3}$ &86.1$_{\pm 1.1}$ & 78.3$_{\pm 0.7}$ & 85.3$_{\pm 0.4}$&75.7$_{\pm 0.4}$\\
    \phantom{ab} 3 heuristics & 9,159 & 91.7$_{\pm 0.5}$  &77.9$_{\pm 1.7}$&51.1$_{\pm 1.4}$&63.9$_{\pm 1.0}$&42.8$_{\pm 0.4}$\\ 
    \phantom{ab} 2 heuristics & 10,621& 87.4 $_{\pm 0.8}$ &70.8$_{\pm 3.3}$&30.1$_{\pm 1.6}$&49.4$_{\pm 1.4}$&19.6$_{\pm 0.5}$\\
    \phantom{ab} 1 heuristic &  2,870& 76.9 $_{\pm 2.3}$ &67.9$_{\pm 3.3}$&25.0$_{\pm 1.4}$&32.1$_{\pm 1.0}$&12.6$_{\pm 0.4}$ \\
    \phantom{ab} 0 heuristic&987& 73.8 $_{\pm 2.3}$ &63.0$_{\pm 3.4}$&33.1$_{\pm 3.1}$&30.8$_{\pm 1.1}$&10.0$_{\pm 1.3}$\\
    \midrule
    \multicolumn{7}{l}{\textit{Subject-verb across object relative}}\\
    \phantom{ab} Overall      & 27,582 &98.9$_{\pm 0.04}$ & 82.0$_{\pm 0.7}$ & 89.1$_{\pm 1.3}$ & 96.7$_{\pm 0.3}$& 86.0$_{\pm 0.6}$  \\\cline{1-7}
    \phantom{ab} 5 heuristics &14,708&99.6$_{\pm 0.05}$ &94.4$_{\pm 0.5}$&97.5$_{\pm 0.4}$&99.1$_{\pm 0.1}$&95.8$_{\pm 0.3}$\\
    \phantom{ab} 4 heuristics &3,799&99.0$_{\pm 0.1}$ &83.1$_{\pm 1.0}$&89.8$_{\pm 1.6}$&96.5$_{\pm 0.4}$&87.3$_{\pm 0.6}$ \\
    \phantom{ab} 3 heuristics &4,189&98.4$_{\pm 0.1}$ &70.8$_{\pm 0.8}$&80.1$_{\pm 3.0}$&93.4$_{\pm 0.4}$&74.9$_{\pm 1.1}$\\
    \phantom{ab} 2 heuristics &3,166&97.7$_{\pm 0.1}$ &54.9$_{\pm 1.4}$&70.1$_{\pm 3.2}$&91.5$_{\pm 0.8}$&64.2$_{\pm 1.3}$\\
    \phantom{ab} 1 heuristics &1,451&96.8$_{\pm 0.1}$ &51.4$_{\pm 1.9}$&73.0$_{\pm 2.3}$&94.1$_{\pm 0.6}$&67.7$_{\pm 1.9}$\\
    \phantom{ab} 0 heuristics &269&94.1$_{\pm 0.5}$ &51.0$_{\pm 2.0}$&67.9$_{\pm 1.9}$&89.3$_{\pm 1.0}$&63.8$_{\pm 1.9}$\\
    
    \bottomrule
  \end{tabular}}
  \caption{Number prediction accuracies before and after the different masking interventions, based on prediction difficulty measured by the number of heuristics. \emph{cue} here refers to the antecedent and its modifiers (determiners and adjectives) in object-past participle agreement and to the subject and its modifiers in subject-verb agreement.}
    \label{res:causal_heuristics}
\end{table*}

 However, probing has a well-known limitation
\cite{belinkov-glass-2019-analysis}: It only brings out a correlation
between the representations and the syntactic information measured by the probe, and does not
tell us whether and how this information is actually involved in
processing the dependency. In this section, we aim to identify which
tokens are actually responsible for providing the number
information. To do so, we design a causal experiment in which we mask
certain words preceding the \emph{target} verb in the sentence to
better determine their role in transformers predictions.

\paragraph{Masking tokens in self-attention computation}

Transformers rely on a self-attention mechanism to build a
contextualized representation for each token by
iteratively defining (as a first
approximation) the token representation as a linear combination of the
representations of the other tokens in the sentence. We propose to
neutralize the contribution of one or more specific tokens in the
construction of the target verb's representation by forcing the weight
of this or these tokens in this linear combination to be zero. This
intervention can be implemented in a straightforward manner by
extending the masking mechanism used in incremental transformers
to prevent a token representation from taking into account future words.

More precisely, we consider the same NA tasks as in Section~\ref{sec:number_agreement_task} but
this time when predicting the target verb (and only at this moment!),
we also mask either the \emph{cue} (and its dependents),\footnote{Masking all
  the \emph{cue} dependents (predicted by an automatic dependency analysis of
  the sentence) allows us to “hide” all tokens with a morphological
  indication of the cue's number such as the determiner and adjectives
  qualifying it.} the relative pronoun \textit{que}, both of these
tokens or all tokens in the \emph{context} except these two tokens.
Table~\ref{ex:mask_que} provides an example of sentence processing
before and after an intervention of masking
\textit{que}. As the intervention occurs only when the target verb is
being predicted, there is no effect on the tokens preceding it. As can
be seen in this example, transformers originally assigned a higher
probability to the correct plural form \textit{accepté·s} than to the
incorrect singular form \textit{accepté}.  After the intervention, the
situation is reversed and the model predicts the (incorrect) singular
form.

This intervention allows us to build counterfactual representations
for both the past participle and the main verb that do not take into
account some tokens in the sentence when computing their
representation, thereby removing any direct access to the information
encoded in the representations of the masked tokens (e.g.\ the number
information encoded in the \emph{cue}).

However, information encoded in these masked tokens can still be taken
into account indirectly: the target verb's representation indeed
relies on the representations of all the preceding tokens in the
sentence, for which the masking mechanism is kept
unchanged---therefore these tokens can still encode relevant information. By
comparing performances on the agreement tasks with and without
intervention, we can evaluate whether the representation of one or
several specific token(s) has a direct impact on the choice of the
target verb form.

\paragraph*{Results} Table~\ref{res:causal_heuristics} reports the
results of our causal interventions on the object-past participle and
subject-verb agreement tasks. Accuracies are broken down by agreement
difficulty.  It appears that, for these two kinds of agreement, the
\emph{cue} (i.e.\ the antecedent group and subject group) is critical
for predicting the corresponding agreement: masking these tokens
strongly degrades transformers prediction performance on the harder
cases (i.e.\ 0, 1 and 2 heuristics subsets). This suggests that
transformers learn representations that are consistent with the French
grammar: the model relies on the same tokens as humans to choose the
correct form of the target past participle and the main
verb.
 
Quite remarkably, the relative pronoun \textit{que} plays a very
different role in determining the form of the target verbs in these
two agreement phenomena: masking the relative pronoun on object-past
participle agreement results in below than chance prediction
accuracies (e.g.\ the accuracy on the \emph{1 heuristic} subset drops
from 76.9\% to 32.1\%), while it hardly impacts the prediction of the
subject-verb agreement: accuracy drops by no more than 6.2 points (for
the \emph{2 heuristics} subset). This suggests that even though the
two agreement phenomena have almost identical surface forms and, as
reported in \textsection\ref{sec:probing}, and the syntactic number
information is encoded in a similar way in both phenomena,
transformers use two distinct agreement mechanisms to resolve the
object-past participle and the subject-verb agreement. This distinction
is consistent with the analysis of theoretical linguistics.

Results reported in Table~\ref{res:causal_heuristics} also show that, in the case of
subject-verb agreement across object relatives, the \emph{context} tokens
(i.e.\ the tokens in the relative clause between the subject and the
verb) have a bigger contribution to the model's decision than the
subject group tokens (i.e.\ the subject and its dependents) with which
the verb agrees. This counter-intuitive observation seems to
confirm the findings of \newcite{ravfogel-etal-2021-counterfactual} that, to predict the subject-verb agreement, the model uses
information about relative boundaries, encoded in its word
representations. To explain our intriguing observation, we hypothesize
that even though the grammatical number is distributed across all
tokens in the \emph{context} segment, the information about relative
boundaries is crucial for the model to determine how to use it to
inflect the main verb, which would explain why the \emph{context} tokens
control the agreement in such an important way. This hypothesis needs
to be confirmed by further experiments.

\section{Discussion \label{sec:disc}}
Our experiments clearly show that transformers are capable of predicting quite well the object-past participle and subject-verb agreements in the case of French object relative clauses. Moreover, even though several confounders exist,  the results consistently indicates that transformers base their predictions mainly on the tokens involved in agreement rules and they apply different strategies despite of the very similar superficial forms of these two linguistic phenomena.

\paragraph{Lexical cues confounder} To evaluate the impact of lexical
information on models predictions, we converted our original evaluation
sets into nonsensical but grammatically well-formed evaluation sets:
following \newcite{gulordava-etal-2018-colorless}, we have created a
\emph{nonce} dataset for each agreement phenomenon by substituting each content word of the original
evaluation sentence with a random word having the same syntactic
category.\footnote{We generated three nonce variants for each original sentence.} We then evaluated our models predictions on number agreement
tasks in this \emph{nonce} setting. The results in
Table~\ref{tab:results_nonce} show only a mild degradation relative to
the \emph{original} setting for the two architectures across the two agreement tasks: a drop of 3.4 percentage points in global performance
for transformers (7.3 for LSTM) in subject-verb agreement, and a drop of
1.5 percentage points for transformers (5.0 for LSTM) in
object-past participle agreement.

This drop in performance is of the same order of magnitude as that
reported by \newcite{gulordava-etal-2018-colorless}, suggesting that
the two models are able to abstract away from the potential lexical
confounds.

\begin{table}[!ht]
   \centering
  \scalebox{.9}{
  \begin{tabular}{llrcc}
    \toprule
    \multicolumn{2}{l}{Evaluation sets} & \makecell{Size \\ {\scriptsize in sentences}} & LSTMs & Transformers \\
    \midrule
    
    \multicolumn{5}{l}{\emph{Object-pp} Original} \\
    & overall    & 68,497 & 82.1 $_{\pm 1.1}$  & 94.6 $_{\pm 0.2}$     \\\cline{2-5}
    \phantom{ab} & singular & 44.599 & 95.4$_{\pm 0.7}$ & 99.2$_{\pm 0.1 }$\\
    \phantom{ab} & plural & 23,898 & 57.2$_{\pm 2.9}$   & 86.2$_{\pm 0.4}$\\
    \midrule
    \multicolumn{5}{l}{\textit{Object-pp} Nonce} \\
                 & overall &68,497*3 & 77.1 $_{\pm 2.3}$  & 93.9 $_{\pm 0.2}$     \\\cline{2-5}
    \phantom{ab} & singular&44.599*3 &90.3$_{\pm 1.3}$&97.5$_{\pm 0.1}$ \\
    \phantom{ab} & plural  &23,898*3 &52.4$_{\pm 5.5}$&87.2$_{\pm 0.6}$ \\
    \midrule
    \multicolumn{5}{l}{\textit{Subject-v} Original} \\
    & overall & 27,582 & 94.3 $_{\pm 0.3}$  & 98.9 $_{\pm 0.04}$     \\\cline{2-5}
    \phantom{ab} & singular & 19,224 & 98.0$_{\pm 0.3}$ & 99.4$_{\pm 0.05 }$\\
    \phantom{ab} & plural & 8,358 & 85.9$_{\pm 1.5}$   & 97.8$_{\pm 0.1}$\\
    \multicolumn{5}{l}{\textit{Subject-v} Nonce} \\
    & overall & 27,582*3 & 87.0 $_{\pm 0.4}$  & 95.5 $_{\pm 0.2}$     \\\cline{2-5}
    \phantom{ab} & singular &19,224*3 &93.7$_{\pm 0.7}$&97.1$_{\pm 0.1}$ \\
    \phantom{ab} & plural &8,358*3 &71.6$_{\pm 2.5}$&91.9$_{\pm 0.4}$ \\
    
    \bottomrule
    \end{tabular}}
  \caption{Accuracy achieved by LSTMs and Transformers on nonce experimental setting, compared to the original setting, by agreement dependency and by target number\label{tab:results_nonce}}
\end{table}

\paragraph{Simplicity confounders} As reported in
Table~\ref{tab:h_accu}, the simple heuristics of
\newcite{li-etal-2021-transformers}, which only rely on surface
information, are able to predict the correct verb form with a very high
accuracy (they even outperform a state-of-the-art LSTM). This
observation, already reported by \newcite{li-etal-2021-transformers}, questions the principle of
using the agreement task as an evidence of syntactic information being
encoded in neural representations. To mitigate this “simplicity”
confounding effect, we have systematically reported our results
according to the “difficulty” of the agreement task and focus, in our
analyses, on the hardest cases that require an abstract representation of the sentence.

It must also be noted that humans also make agreement errors
\cite{BOCK199145}, these sequential statistical heuristics and the
method of sampling evaluation sets based on heuristics may therefore
serve as a source of hypotheses for experiments assessing human
syntactic abilities.

\begin{table}
  \scalebox{.85}{
  \begin{tabular}{lcc}
    \toprule
    Heuristics & object-pp & subject-verb \\
    \midrule
    (1) First noun        & 69.5 & 83.7\\
    (2) Most recent noun         & 88.6 &77.5\\
    (3) Most recent token        & 60.3 &66.9\\
    (4) Majority number   & 70.0 &75.9    \\
    (5) Noun before “que” & 95.7 & 91.6 \\
    \bottomrule
  \end{tabular}}
\caption{Accuracy (\%) achieved by the 5 surface heuristics considered in
  this work on the long-distance agreement task. \label{tab:h_accu}}
\centering
\end{table}

\paragraph{Frequency bias and imbalanced distribution confounders} Given the fact that in written French, singular verbs (3rd person) occur five to ten times as often as their plural counterparts \cite{aagren2013input}, do our models' performance depend on the number of the \emph{target} verb? As  table~\ref{tab:results_nonce} shows, in the \emph{original} setting, across the two agreement dependencies, both models perform consistently better on the singular condition
than on the plural condition, suggesting a bias towards singular verbs in both models.  This observation is in line with the conclusions of \newcite{wei-etal-2021-frequency}, who found that more frequent forms are more likely to be better predicted in number agreement tasks. At the same time, the results of our intervention experiments reveals the similar pattern: the masking interventions lead to greater degradation for the plural condition across the board. The remarkable different contribution of ``que'' persists, even though the plural condition is mainly responsible for the below than chance accuracy after masking ``que'' intervention in object-past participle agreement. 

A further analysis of our evaluation datasets reveals that the relation between the class distribution and the task difficulty (measured by the count of heuristics in \textsection\ref{sec:number_agreement_task}) shows a very similar pattern across both agreement dependencies, with the singular class dominating in the easiest cases---\emph{5 heuristic} subset (Obj-pp: 94\%, Subj-v: 91\%) and the plural class dominating in the most difficult cases---\emph{0 heuristic} subset (Obj-pp: 99\%, Subj-v: 96\%). 

Therefore, this asymmetry in singular and plural condition, in terms of prediction accuracy on number agreement tasks,  may be either due to the higher frequency of singular verbs in the language, or to the frequency imbalances in verb number across syntactic constructions, or to the different ways the model encodes singularity and plurality as suggested by \newcite{jumelet-etal-2019-analysing}.
In future work, these hypotheses could be tested by artificially manipulating the relative frequency of singular and plural nouns in various constructions in the training corpus.  

\section{Related Work \label{sec:rw}}

\subsection{Understanding the Inner Working of Neural Networks}

Three kinds of approaches have been used in the literature to analyze
the inner working of neural
networks~\cite{rogers-etal-2020-primer,belinkov-etal-2020-interpretability}.
Our analyses, based on a combination of these existing methods, bridge
the gap between the representations and behavioral approaches, and provide a framework for
systematically measuring the causal factors underlying the model's
behavior in order to evaluate a precise hypothesis about the abstract
structure.

\paragraph{Linguistic Probes} Probing \cite{alain17understanding,hupkes2018visualisation} consists in
training a supervised classifier to predict linguistic properties from models' internal representations; achieving high accuracy in this task implies that these
properties were encoded in the representation. Substantial prior work (see \newcite{belinkov-2022-probing} for a review) has used this approach  to assess neural networks linguistic capacity, finding that NLMs encode a variety of information about, among others, grammatical number, part of speech and syntactic role \cite{giulianelli-etal-2018-hood, tenney2018what,jawahar-etal-2019-bert}. 

Many recent studies, such as \cite{hewitt-liang-2019-designing} or
\cite{pimentel-etal-2020-information}, have criticized the theoretical
foundations of probing, in particular by raising issues related to the
capacity of the classifiers used for probing representations. There
is an ongoing debate as to whether the choice of a linear classifier
(as we did in our experiments) is the right decision, in particular
when comparing the capacity of two neural networks to capture a given
linguistic phenomenon~\cite{pimentel-etal-2020-pareto}. As we are only
interested in detecting whether one linguistic property is encoded and consider
a single model in our experiments, the choice of a linear classifier
is fully justified.

\paragraph{Causal Approaches} These approaches rely on interventions
that modify parts of the neural networks and analyze the
impact/consequences on models
output~\cite{vig20investigating,ravfogel-etal-2021-counterfactual,lasri-etal-2022-probing}. They allow to avoid one of the main pitfall of linguistic probes namely
that probes can only detect correlations: as shown by \newcite{vanmassenhove17investigating}, information encoded in the representation (revealed by the probe) is not necessarily used by the model.
\newcite{giulianelli-etal-2018-hood}, one of the first work to use
this approach, combined the probing and causal intervention by using a
trained probe to modify NN's representations, which allows them to
improve the model's agreement
prediction. \newcite{finlayson-etal-2021-causal} performed controlled
interventions on input sentences to analyse NLM's syntactic agreement
mechanism, showing that LMs rely on similar sets of neurons when processing similar syntactic structure. Our intervention
described in \textsection\ref{sec:causal}, is most closely related to
\newcite{elazar-etal-2021-amnesic}, who proposed to erase the
“part-of-speech property” encoded in a model representation to measure
how important this information is for word prediction. Our approach
differs from much prior work in causal analysis in two aspects: i) We
perform intervention on the attention mechanism rather than on the
input sentences. ii) We test linguistically motivated hypotheses
through two phenomena with very similar surface forms, which provides
more fine-grained insights and can be easily extended to other
linguistic hypotheses.

\paragraph{Challenge Sets} This approach seeks to assess the
linguistic competence of a model on examples carefully selected to
exhibit a particular language characteristic
\cite{isabelle-etal-2017-challenge}. For instance,
\newcite{linzen-etal-2016-assessing}, an iconic example of this approach
on which our work is based, studies the ability of a neural network to learn syntax-sensitive dependencies
and in particular the agreement between the subject and the
verb. \newcite{warstadt-etal-2019-neural} and \newcite{warstadt-etal-2020-blimp} 
 introduce challenges sets covering
a wide range of grammatical phenomena in English. To the best of our
knowledge, most of existing challenge sets are for English; our work perfectly illustrates the benefit of extending this approach to non-English languages: considering French, a morpho-syntactically richer language, allows us to evaluate neural networks in a more fine-grained way and to gain additional insights into their inner working.

\subsection{Number Agreement Task}

The prediction of subject-verb agreement has long been identified as a
way to assess the capacity of neural networks to track syntactic
dependencies: it was already used by \newcite{elman89representation},
one of the first work on the analysis of the inner working of neural
networks. Revitalized by the seminal work of
\newcite{linzen-etal-2016-assessing}, this task has been used in a
tremendous number of studies to bring to light the ability of neural
networks to capture abstract information.

Building on these results, a first line of research has tried to
provide further evidence regarding neural networks capacity to build an
abstract representation of sentences either by generalizing them to
other languages \cite{lakretz2021mechanisms}, or to other model
\cite{goldberg19assessing} or by identifying possible confounds such
as lexical co-occurrences \cite{gulordava-etal-2018-colorless} or
unbalanced datasets \cite{li-etal-2021-transformers}.

Another interesting line of research aims at generalizing the
conclusions of \newcite{linzen-etal-2016-assessing} to other
constructions. For instance, \newcite{marvin-linzen-2018-targeted}
evaluated the grammaticality judgement of a neural network on a wide array of linguistic phenomena, using
carefully designed templates to generate
sentences and \newcite{li-etal-2021-transformers} investigated the
object-verb agreement in French.

A few studies have questioned the conclusions drawn by all these
works. For instance, \newcite{newman-etal-2021-refining} questioned the
use of the 0/1 loss to evaluate the number agreement task, arguing that
this evaluation does not take into account the probability
distributions over the vocabulary and, when not restricted to choose
between two verb forms, the model might actually give more weights to
verbs with the wrong number. \newcite{lasri-etal-2022-bert} observed
that the capacity of a model to predict the correct form of the verb
decreased with the sentence complexity, an observation confirmed by
the results reported in Section~\ref{subsec:number_pred_results} or reported by
\newcite{li-etal-2021-transformers}. Despite these potential confounds, our results align with these authors, confirming that the drop in performance is not sufficient to invalidate the conclusions of \newcite{linzen-etal-2016-assessing}.

\section{Conclusion \label{sec:conclu}}

This work addresses the question of the capacity of transformers to
uncover abstract representation of sentences rather than to merely
capture surface patterns. To this end, we investigate the mechanisms
implemented by this model to predict two kinds of agreement
in French: the subject-verb and the object-past participle
agreements. Even though these two agreements involve very similar word
sequences, their linguistic analyses are fundamentally different.

The first set of experiments we reported show that transformers are
indeed able to predict these two kinds of agreement and, by comparing
their predictions with those of simple surface heuristics, we highlight their
ability to capture syntactic information. In a second set of
experiments, we used probing and counterfactual analysis methods
to provide evidence that transformers are actually succeeding for good
reasons: our analyses reveal that \textit{i)} the incremental transformer bases its
predictions on cues that are linguistically
motivated and consistent with the French grammar, \textit{ii)} the
abstract structure uncovered by transformers reflects the distinction
made in the theoretical modeling of these two agreements.

Our work is a first step towards a better understanding of the inner
representations of NNLMs. Designing new probes, supported by
causal analysis and involving a wider range of languages, could improve
our understanding of such models. In particular, our observation about
the linguistically motivated distribution of syntactic information in
transformers' representations could be extended to other linguistic
phenomenon and languages.

\section*{Acknowledgments}
We sincerely thank the reviewers and action editors for their careful
reviews and insightful comments, which are of great help in improving
the manuscript. This work was granted access to the HPC resources of
French Institute for Development and Resources in Intensive Scientific
Computing (IDRIS) under the allocation 2020-AD011012282 and
2021-AD011012408 made by GENCI. We would also like to gratefully acknowledge support from the Labex EFL (ANR-10-LABX-0083).

\vspace{5mm}

\appendix
\section{Evaluation Datasets}\label{sec:descrip_stats}
Although in standard French, normative grammars indicate object-past participle agreement under wh-movement as obligatory, it in fact appears to be optional in colloquial French where the past participle is often produced in its default singular, masculine form~\cite{belletti2017past}. Therefore, to determine to which proportion the sentences in the training data of language models respect the normative past participle agreement rule, we analysed the training data with the same automatic extraction procedure used for constructing evaluation datasets (section \ref{sec:long-distance}). We found that in 10,377 sentences (0.35\% of training data) containing the target filler-gap dependency, 10\% of them don't respect the agreement rule. 

\section{Language Models}\label{sec:LM} 

\paragraph{Hyperparameters and perplexities} The results reported in the paper are averaged over five best models in terms of the validation perplexity. 

The total parameters of the LSTM models are 47,900,241. As described in the appendix of \newcite{li-etal-2021-transformers}, the model of batch size 64, with dropout 0.1 and learning rate 0.0001 achieved the lowest perplexity scores: 37.1, we then trained four LSTM models with the same combination of hyperparameters, the resulting perplexities are: 36.8, 36.8, 36.9 and 37.0. 

The total parameters of the transformer models are 126,674,513. We explored the initial learning rate of 0.01 and 0.02, the dropout rate of 0.1, 0.2, 0.3, 0.4 resulting in a total of 8 combinations. The model with learning rate 0.02 and dropout rate 0.2 achieved the lowest perplexity---27.0, we trained another four transformer models with these hyperparameters and the resulting perplexities are: 26.8, 27.0, 27.1 and 27.2. Training was performed with stochastic gradient descent and used the same hyperparameters described in appendix of \newcite{li-etal-2021-transformers}: the initial learning rate was fixed to 0.02 and we used a cosine scheduling on 90 epochs without annealing. The first epoch was dedicated to warmup with a linear incremental schedule for the learning rate. Batches of size 64 ran in parallel on 8 GPUs except for warm-up, where the size was fixed to 8. 

\section{Detailed Number Predictions Results}\label{sec:graph}
We also report the detailed scores of Figure~\ref{tab:original-heuristics_graph} in table~\ref{tab:original-heuristics}.

\begin{table}[h]
  \centering
  \scalebox{.8}{
  \begin{tabular}{llrcc}
    \toprule
    \multicolumn{2}{l}{Constructions} & \makecell{Size \\ {\scriptsize (in sentences)}} & LSTMs & Transformers \\
    \midrule
    \multicolumn{5}{l}{\textit{Object past participle}} \\
    \phantom{ab} & overall & 68,497 & 82.1$_{\pm 1.1}$   & 94.6 $_{\pm 0.2}$ \\\cline{2-5}
    \phantom{ab} & 5 heuristics &32,149 &95.3 $_{\pm 0.6}$ & 99.2 $_{\pm 0.1}$ \\
    \phantom{ab} & 4 heuristics &12,711 &85.9 $_{\pm 1.0}$ & 96.5 $_{\pm 0.1}$ \\
    \phantom{ab} & 3 heuristics & 9,159 & 71.9 $_{\pm 1.6}$ & 91.6 $_{\pm 0.4}$\\
    \phantom{ab} & 2 heuristics & 10,621 & 62.2 $_{\pm 2.4}$& 87.6 $_{\pm 0.4}$\\    
    \phantom{ab} & 1 heuristic & 2,870&37.4 $_{\pm 4.1}$& 77.9 $_{\pm 0.8}$\\
    \phantom{ab} & 0 heuristic & 987&40.2 $_{\pm 2.7}$ & 76.1 $_{\pm 1.0}$\\
    \midrule
    \multicolumn{4}{l}{\textit{Subject-verb across object relative clause}} \\
     & overall & 27,582 &94.3$_{\pm 0.3}$& 98.9$_{\pm 0.04}$    \\\cline{2-5}
     & 5 heuristics &14,708 &98.6$_{\pm 0.1}$&99.6$_{\pm 0.05}$\\
     & 4 heuristics &3,799&95.2$_{\pm 0.5}$&99.0$_{\pm 0.1}$ \\
     & 3 heuristics &4,189 &91.3$_{\pm 0.8}$&98.4$_{\pm 0.1}$ \\
     & 2 heuristics & 3,166 &84.8$_{\pm 1.0}$&97.7$_{\pm 0.1}$ \\    
     & 1 heuristic &1,451  &81.2$_{\pm 1.8}$&96.8$_{\pm 0.1}$ \\
     & 0 heuristic  &269 &74.7$_{\pm 2.2}$&94.1$_{\pm 0.5}$ \\
    \bottomrule
  \end{tabular}}
\caption{Accuracies(\%) achieved by LSTMs and transformers as a
  function of the agreement prediction difficulty, on the object-pp
  and the subject-verb agreement tasks \label{tab:original-heuristics}
}
\end{table}

\bibliography{tacl2021,anthology,custom}
\bibliographystyle{acl_natbib}


\end{document}